\documentclass[conference]{IEEEtran}
\IEEEoverridecommandlockouts
\usepackage{cite}
\usepackage{amsmath,amssymb,amsfonts}
\usepackage{algorithmic}
\usepackage{graphicx}
\usepackage{textcomp}
\usepackage{booktabs}
\usepackage{lipsum}

\usepackage{xcolor}
\def\BibTeX{{\rm B\kern-.05em{\sc i\kern-.025em b}\kern-.08em
    T\kern-.1667em\lower.7ex\hbox{E}\kern-.125emX}}
\begin{document}

\title{Emotion Recognition for Healthcare Surveillance Systems Using Neural Networks: A Survey}
\author{\IEEEauthorblockN{Marwan Dhuheir, Abdullatif Albaseer, Emna Baccour, Aiman Erbad, Mohamed Abdallah, and~Mounir Hamdi}
\IEEEauthorblockA{Division of Information and Computing Technology, College of Science and Engineering,
\\Hamad Bin Khalifa University, Qatar Foundation, Doha, Qatar \\
\{Marwan, amalbaseer, EBaccourEpBesaid, AErbad, moabdallah, mhamdi\}@hbku.edu.qa}
}
\maketitle

\begin{abstract}
Recognizing the patient's emotions using deep learning techniques has attracted significant attention recently due to technological advancements. 
Automatically identifying the emotions can help build smart healthcare centers that can detect depression and stress among the patients in order to start the medication early. Using advanced technology to identify emotions is one of the most exciting topics as it defines the relationships between humans and machines. Machines learned how to predict emotions by adopting various methods. In this survey, we present recent research in the field of using neural networks to recognize emotions. We focus on studying emotions' recognition from speech, facial expressions, and audio-visual input and show the different techniques of deploying these algorithms in the real world. These three emotion recognition techniques can be used as a surveillance system in healthcare centers to monitor patients.  We conclude the survey with a presentation of the challenges and the related future work to provide an insight into the applications of using emotion recognition.
\end{abstract}


\section{Introduction}

The advancement of deep learning technologies brings more attention to deployment scenarios in smart health systems\cite{8674240,8450511,9076126}. The health industry uses many approaches based on machine learning, such as remote disease diagnosis, surveillance system in healthcare and elderly care centers, etc., to recognize patients' emotions. These systems are used for early emotion recognition to introduce prompt interventions reducing symptoms of depression and stress. In this survey, we present the recent work of using three different techniques to recognize emotions which are speech, facial and audio-visual. We focus on using a deep neural networks to identify patients’ emotions. These methods can be used as a surveillance system and capture images, videos, and speech using different tools such as cameras and microphones.
The areas that use patient's emotion recognition are wide, and its applications include many necessary daily life uses such as in safe driving, monitoring mental health, social security, and so on. Many surveys cover this topic with deep details from different perspectives, such as in \cite{egger2019emotion,noroozi2018survey,mehta2018facial,noroozi2017audio,li2020deep,dehghan2017dager,akccay2020speech}. The authors studied emotion recognition by focusing on body gesture, speech expressions, and audio-visual expressions. The authors of these surveys focused on multi-modal approaches that study either face or speech with body gestures to enhance the emotion recognition. The survey studies in \cite{mehta2018facial}\cite{li2020deep} focus on studying emotion recognition by using facial emotion, and they used a device called Microsoft HoloLens (MHL) to observe the emotions in Augmented Reality (AR). They used the device as a sensor in the experiment to recognize emotions. Then, they compared their method with a method that recognizes emotion by using a webcam. The experiment concluded that using MHL with AR gave better accuracy than using a webcam.

Studying Patients Emotion recognition for better health system has become necessary in the last decade as it helps in many fields, and one of these fields is the medical sector. It helps doctors recognize the patients' psychological problems and, hence, start the medication early\cite{egger2019emotion}. Many hospitals worldwide have begun incorporating AI in medicating patients, and many researchers are focusing on studying neural networks to recognize the patient's emotions. This survey presents one common AI technique to recognize emotions by using three different modalities: speech, facial, and audio-visual methods. We present the common techniques used to recognize emotions to give the readers a general overview of using neural networks in the medical sector.

Our contribution in this survey is to study patients emotion recognition techniques which is considered a key to enhancing patients healthcare. Although many techniques are used for emotion recognition, such as recognizing emotions by using Electroencephalography (EEG), Electrocardiography (ECG), respiration, gesture, etc., we focus on the three methods captured by surveillance systems using cameras and microphones. The study focuses on three essential stages to make the final recognition decision: pre-processing, feature selection and extraction, and classification. We highlight the recent techniques and scenarios used in each stage. 

The paper is organized as follows: Section II presents the common datasets, and section III illustrates speech emotion recognition. Section IV shows facial emotion recognition. Section V presents the Audio-visual emotion recognition. Finally, section VI presents the conclusion and future work. 

\section{Databases and Test preparation}
This section presents the commonly used datasets in recognizing emotions. To effectively design an emotion recognition system, it is crucial to have training data that comprises many different populations and environments. These datasets are utilized for training the suggested methods and approaches; therefore, they should be chosen carefully to conduct suitable experiments. In Table~\ref{tab:data_summary}, we summarize and describe these datasets.

\begin{table*}[!ht]
\label{mainref}
    \centering
    \begin{tabular}{|c|p{2cm}|p{2cm}|p{4cm}|p{7cm}|}\hline
        Reference & Dataset & Type & calsses & Description \\\hline
       \cite{noroozi2018survey} & Extended Cohn-Kanade \textbf{(CK+)} &  laboratory-controlled & anger, disgust, fear, happiness, sadness, and surprise  & consisting of 593 sequences from 123 subjects\\\hline
       \cite{noroozi2018survey} & M\&M Initiative \textbf{(MMI)} &  laboratory-controlled & anger, disgust, fear, happiness, sadness, and surprise  & consisting of more than 1500 samples of image sequences and static images of faces\\\hline
       \cite{akccay2020speech} & \textbf{Oulu-CASIA} &  laboratory-controlled & happiness, sadness, surprise, anger, fear, disgust & containing  2880  image  sequences  that  were  collected  from  80  subjects.\\\hline
     \cite{akccay2020speech} & \textbf{(JAFFE)} &  Scientific research & (happiness, neutral, sadness, anger, surprise, disgust, fear) & comprising of 213 samples of posed expressions that were taken from Japanese females\\\hline
    \cite{li2020deep} & \textbf{FER2013} &  open-source dataset & happiness, neutral, sadness, anger, surprise, disgust, fear & comprising of 35,685 samples of 48x48 pixel grayscale images for facial expression.\\\hline
    \cite{akccay2020speech} & \textbf{AFEW} and \textbf{SFEW} &  open-source dataset & happiness, neutral, sadness, anger, surprise, disgust, fear & consists of video clips that were gathered from various movies with unconstrained expressions, different head poses, occlusions, and illumination\\\hline
    \cite{akccay2020speech} & \textbf{Multi-PIE} &  open-source dataset & happiness, sadness, anger, surprise, disgust, fear & These images were collected from 337 subjects under 15 viewpoints, and the illumination was 19\\\hline
    \cite{noroozi2018survey} & \textbf{ BU-3DFE and BU-4DFE} &  Research-oriented dataset & happiness, sadness, anger, surprise, disgust, fear & consisting of 606 facial expression sequences from 100 people and the around 60,600 frame models. \\\hline
    
     \cite{mirsamadi2017automatic} & \textbf{EmotioNet} &  Public dataset & happiness, sadness, anger, surprise, disgust, fear & consisting  of one million facial expression images gathered from the Internet\\\hline
     \cite{akccay2020speech} &\textbf{RAF-DB} &  Public dataset & happiness, sadness, anger, surprise, disgust, fear & comprising of 29,672 diverse facial images that were downloaded from the Internet\\\hline
     \cite{li2020deep} & \textbf{AffectNet} & Public dataset & happiness, sadness, anger, surprise, disgust, fear & consisting of more than one million images gathered from the Internet\\\hline
     \cite{jain2019extended} & \textbf{EMO-DB}  &  Research-oriented dataset & anger, joy, sadness, neutral, boredom, disgust, and fear & containing 535 emotional expressions\\\hline
     \cite{noroozi2017audio} & \textbf{RML}  &  Research-oriented dataset & anger, disgust, fear, joy, sadness, and surprise & consisting  of 720 utterance expressions with eight subjects\\\hline
     \cite{huang2019speech} &\textbf{eNTERFACE05}  &  audio-visual emotion dataset & anger, disgust, fear, joy, sadness, and surprise & consisting of 1290 utterances \\\hline
    \cite{noroozi2017audio} &\textbf{BAUM-1s}  &  audio-visual emotion dataset & joy, anger, sadness, disgust, fear, surprise, boredom, contempt, unsure, thinking, concentrating, and bothered  & consisting of 1222 utterances gathered from 31 Turkish subjects\\\hline 
    \end{tabular}
    
    \caption{Summary of the main utilized Datasets.}
    \label{tab:data_summary}
\end{table*}

\section{Speech Emotion Recognition}

Speech is the communication medium between humans. Many researchers use machines to interact with humans and to extract their feelings. However, it requires an extensive effort to make this interaction natural between humans and machines. In this section, we present the recent work in using machines to predict emotions, focusing on the different methods using neural networks. Using neural networks gives an advantage in terms of the efficiency to extract emotions from speech due to the automatic feature selection, which is challenging in traditional speech emotion recognition (SER) techniques.

Recently, speech emotion recognition is becoming an attractive approach. New techniques deal with the complexity in extracting emotions and are affected by different factors such as age, gender \cite{dehghan2017dager}, and the difficulty to process large data sets. The study in \cite{mirsamadi2017automatic} explored speech emotion recognition using recurrent neural networks (RNN) with a  focus on local attention. They used DL to learn short-time frame-level acoustic features and a suitable aggregation of these features. They used local attention because it can focus on particular regions of the more emotionally salient speech signal. This method is more accurate than the traditional SVM-based speech emotion recognition (SER) that uses fixed designed features. Furthermore, the speech emotion recognition is studied in \cite{lim2016speech} \cite{badshah2017speech}\cite{tzirakis2018end}\cite{basu2017emotion} in which the authors focused on proposing an SER algorithm based on concatenated convolutional neural network (CNN) and RNN without using any hand-crafted extracting method.
Authors in \cite{tzinis2017segment} studied speech emotion recognition by extracting the statistical features over segments of speech. The segments are extracted according to the matching of a couple of words. They tested it on Interactive Emotional Motion Capture (IEMOCAP) database. Learning utterance-level representation for speech emotion recognition was covered in \cite{wang2017learning}, and they focused on encoding each utterance into a vector by pooing the activation of the final hidden layer. They formulated an optimization problem to minimize the utterance-level target. To differentiate between verbal and nonverbal speech sounds in real-life conversations, the study in \cite{huang2019speech} used Prosodic Phrase (PPh) auto-tagger to extract the verbal and nonverbal segments, and the result showed that the nonverbal speech intervals gave an excellent performance and the sound feature ability to identify emotion recognition.

\subsection{Preprocessing}

After collecting data to be classified, it goes to the preprocessing step to prepare and mitigate the effects of noise. The input data is corrupted by noise that needs to be removed; otherwise, feature selection and extraction will not be sufficient enough for the classification \cite{9223606}. Several preprocessing techniques are used for feature extraction. Some methods do feature normalization so speakers and recordings variations do not influence the process of recognizing emotions using speech \cite{castillo2016software}.

Depending on the type of input data, a suitable approach can be applied. For example, voice input data are preprocessed to extract the data segments using the vocal cords' quasi-periodic vibration. In contrast, unvoiced input data are preprocessed by using turbulent airflow \cite{vandermosten2020brain}. Other methods commonly used are framing, windowing, and normalization, in which the choice of choosing the suitable method depends on the type of the input voice data \cite{akccay2020speech}. Many standard preprocessing techniques are used as noise reduction of the input data, like minimum mean square error (MMSE) and log-spectral amplitude MMSE (logMMSE) \cite{tzirakis2018end},\cite{basu2017emotion},\cite{wang2017learning}. Other efficient methods use sampling and frame operations to obtain a set of labeled samples \cite{zhou2016deep}.

\subsection{Feature Selection and extraction}

The speech signal is continuous by nature and carries information, and it contains emotions. Hence, according to the followed feature approach, global or local features can be selected accordingly. Global features, known as long-term or supra-segmental features, express the gross statistics such as mean, minimum, maximum values, and standard deviation. On the other hand, local features, known as short-term or segmental features, express the temporal features, in which the main goal is to approximate the stationary state\cite{huang2019speech}\cite{lim2016speech}. As emotional features are distributed in a non-uniformly manner over all speech signals, the stationary states become crucial to be adopted \cite{akccay2020speech}. Table ~\ref{tab:features} presents details about the common features used in SER.

\begin{table*}[!ht]
\label{mainref}
    \centering
    \begin{tabular}{|c|p{1.8cm}|p{14cm}|}\hline
 Reference & Feature & Description \\ [0.5ex] 
 \hline
 \cite{busso2009analysis} & Prosody & Features that human perceives like intonation and rhythm. SER uses mostly frames duration, intensity, and contour of fundamental frequency F0 for prosody features. Its frame duration typically ranges between 30-100 ms.\\ 
 \hline
 \cite{wu2011automatic} & Spectral & It aims to obtain the energy content of the available frequency bands in the speech signals. Spectral features commonly used are formant, cepstral, MFCC, linear predictive cepstral coefficient (LPCC), and perceptual linear prediction (PLP).\\
 \hline
 \cite{kim2015kinematic} & Voice quality & It is obtained by the physical characteristics of the vocal tracts. The variations of the speech signals like jitter, shimmer and harmonics are defined as the constructions of voice quality features. Its duration is less than 10 ms, hence it is called sub-segmental level features. \\
 \hline
 \cite{zao2014time} & Non-linear & It is produced when vocal cords exert non-linear pressures, hence it cannot be represented by using traditional features methods. Nonlinear dynamic (NLD) was introduced to represent the features\\
 \hline
 \cite{badshah2019deep} & Deep-learning-based & Deep learning algorithms can be used to learn both low-level and high-level features hierarchically. The low-level descriptors (LLD) algorithms can be applied directly to deep learning algorithms. \\
\hline
\cite{yenigalla2018speech} & Non- linguistics vocalization & It contains speech disfluencies like laughter, breathing, crying, and different breaks. These features are important for SER and can be recognized by using an automatic speech recognition engine. \\
 \hline
\end{tabular}
    
    \caption{features used in SER.}
    \label{tab:features}
\end{table*}
\setlength{\textfloatsep}{0pt}
\subsection{Classifiers}

The classification of SER systems depends on the utterance of emotions. The classifiers can be divided into two parts. One followed the traditional classifiers such as the Hidden Markov Model (HMM) and Artiﬁcial Neural Networks (ANN). The second one uses deep learning algorithms. However, nowadays, most of the classification processes are done using deep neural networks as they can do feature selection and extraction at the same time \cite{tzinis2017segment}. 
Table~\ref{tab:classifiers} presents the common classifiers that are used in recognizing emotions by using speech.

\begin{table*}[!ht]
\label{mainref}
    \centering
    \begin{tabular}{|c|p{2.3cm}|p{13cm}|}\hline
    \multicolumn{3}{|c|}{Traditional based classifiers} \\
    \hline
 Reference & Classifier & Description \\ [0.5ex] 
 \hline
 \cite{ntalampiras2011modeling} & Hidden Markov Model (HMM) & In this model, the current state depends on previous state. This model uses little data to recognize speech emotions from contextual information. Its strong side is in natural databases.\\ 
 \hline
 \cite{busso2009analysis} & Linear Discriminant Analysis (LDA) & To classify the input speech data, it uses dimensional reduction of input data, hence decreases the computational load.\\
 \hline
 \cite{lee2011emotion} & Singular Vector Machine (SVM) & It achieves better in case of small databases and high dimension features in which these two features are common in SER. It also cares about both testing and training data.\\
 \hline
 \cite{lugger2007relevance} & k-Nearest Neighbor (k-NN) & It deals more with nonlinear feature inputs to create the relations. The disadvantages of this classifier, distance and k calculations are important.\\
 \hline
 \cite{gharsalli2016feature} & Ensemble  Classiﬁers (EC) & It minimizes the variances and decreases the over-fitting which are crucial in SER. If the features are correlated, this method cannot work well.\\ 
\hline
\cite{ververidis2005emotional} & Gaussian Mixture Model (GMM) & It can work well when it is combined with discriminate classifiers like SVM because it can generate and learn the hidden features of speech emotion. \\

 \hline
    \multicolumn{3}{|c|}{Deep learning based classifiers} \\
    \hline 
     \cite{abdelwahab2017ensemble} & Artiﬁcial Neural Networks (ANN) & This classifier achieves good results for nonlinear emotional features. Its latency is very short to predict the features, hence it is effective for applications that are sensitive to time.\\
 \hline
 
 \cite{mao2014learning} & CNN & It has ability to decrease the signal processing, automatic learning of discriminative and global emotional features.\\
 \hline
 \cite{mirsamadi2017automatic} & LSTM & It has the ability to process the long contextual information and the long variance input utterance features.\\ 
\hline
\cite{deng2017universum} & Auto-Encoder Neural Network (AEN) & It has the ability to work in mismatched environments, it can learn features in low dimensional spaces, and nonlinear features.\\[1ex] 
 \hline
\end{tabular}
    
    \caption{Different Classifiers in SER.}
    \label{tab:classifiers}
\end{table*}
\setlength{\textfloatsep}{0pt}  
\section{Facial Emotion Recognition}

The second emotion recognition is to use DNN for facial emotion recognition. It is helpful as it depends on the images captured by camera in the healthcare units and process them. Authors in \cite{mehta2018facial} \cite{jain2019extended} used batch normalization to improve both generalization and optimization. The experiment was conducted on the Extended Cohn-Kanada (CK+) and Japanese Female Facial expression (JAFFE) datasets. The result showed that the fully convolutional network (FCN) and residual block cloud improve the system efficiency.


\subsection{preprocessing}

Facial expressions contain many irrelevant variations such as various background, image illumination, and body poses, therefore, preprocessing data input is important. DNNs are used to learn many features and propose preprocessing to align and normalize the data captured from the faces. Table~\ref{tab:Preprocessing} presents the common preprocessing techniques that are used in recognizing emotions by using images.

Authors in \cite{jain2019extended}\cite{jain2018hybrid} used the normalization method for preprocessing input data to specify the face in the picture and identify the points of interest before passing the data to feature extraction. Moreover, the authors in \cite{jain2019extended} further divided the image normalization into two subtasks, subtracted local contrast and divisive local contrast, and decreased the mismatch in the image content. In \cite{li2020deep}, the authors used the face alignment method to process the input image in which they used an affine transformation specified by the centers of the eyes and the center of the mouth. The output data are further preprocessed by the normalization method. 

\begin{table*}[!ht]
\label{mainref}
    \centering
    \begin{tabular}{|c|p{2.5cm}|p{13cm}|}\hline
 Reference & Preprocessing method & Description \\ [0.5ex] 
 \hline
 \cite{ji2012automatic} & Normalization & It uses median filters to reduce the illumination and variations of the input images and improve the image quality.\\ 
 \hline
 \cite{noh2007feature} & Localization & This preprocessing method uses the algorithm of Viola-Jones to recognize the input image. To detect the faces' size and location, Adaboost learning algorithm and haar lhaar-likeres algorithms are used.\\
 \hline
 \cite{li2020deep} & Face Alignment & It is used to remove the background and the areas that do not contain the face. To do that, the Viola-Jones (V-J) face detector is used for face detection because it is robust.\\
 \hline
 \cite{li2020deep} & Data augmentation & There are two types of data-augmentation approaches: 1) on-the-fly data augmentation and 2) offline data augmentation.\\ 
\hline
\cite{uccar2016new} & Histogram Equalization Method & It is used to overcome the variations in the image illuminations. This method is used to improve the contrast of the images and improve the face images' lighting. \\[1ex] 
 \hline
\end{tabular}
    
    \caption{Preprocessing in FER.}
    \label{tab:Preprocessing}
\end{table*}
\setlength{\textfloatsep}{0pt}
\subsection{Feature Selection}

Deep learning is one effective technique to capture the features from the images. It can capture high-level details through hierarchical structures of many non-linear transformations and representations. The stage of feature selection includes selecting the training set for making them ready for machine learning algorithms. It focuses on choosing a suitable prediction for the learning system. This step helps to enhance the rate of prediction and enhance efficiency. Several tools are useful such as Weka and sci-kit learn which contain inbuilt tools for effective automated selection of features. 

The study in \cite{jain2019extended} used DNN to extract the feature from a set of images in which they used several datasets such as CK+, JAFFE, and Cohn-Kanade. The result shows a performance improvement. The authors in \cite{li2020deep} used different CNN techniques to extract features such as C3D for spatiotemporal extraction.

\subsection{Feature classification}

It is the last stage in FER's system, which gives the final decision of the detected emotion such as sadness, anger, disgusting, etc. The widely used classification methods are SVM, Nearest Neighbor (NN), Softmax, Deep Neural Forest (NFs). The input extracted data uses either a particular face action or a specific face emotion for classification; however, the latter is commonly used \cite{li2020deep}.

Authors in \cite{jain2019extended} used Softmax classifier to classify the incoming data into six demotions. They used a single DNN that contains both convolutional layers and residual blocks to achieve higher accuracy and train deeper layers. Their model showed better performance and better accuracy than the state-of-the-art methods. Other studies \cite{li2020deep}\cite{7752782} used the SVM classifier. They first applied SVM with radial basis function (RBF) and optimized the output data using the grid search optimization method. Table ~\ref{tab:Classifiers in FER} gives the most common classifiers that are used in FER.

\begin{table*}[!ht]
\label{mainref}
    \centering
    \begin{tabular}{|c|p{2cm}|p{13cm}|}\hline
 Reference & Classifier & Description \\ [0.5ex] 
 \hline
 \cite{li2020deep} & Softmax loss & To minimizes the cross-entropy between the calculated class likelihood and the ground-truth distribution. \\ 
 \hline  
 \cite{kontschieder2015deep} & Deep Neural Forest (NFs) & It uses NFs instead of softmax loss, and they achieved a similar result for recognizing emotions from faces images.\\
 \hline
 \cite{zhang2014random} & Support Vector Machine (SVM) & It is a supervised machine learning technique that uses four types of kernels to improve the performance of classification. These four kernels are linear, polynomial, Radial Basis Function (RBF) and sigmoid, and they work together to improve the performance.\\
 \hline
 \cite{noh2007feature} & ID3 Decision Tree (DT) & This classifier is a rule based which uses the decision tree to extract the rules. Least Boolean evaluation is used to execute the classification.\\ 
\hline
\cite{vickers2017animal} & (MFFNN) & This classifier utilizes three different layers which are input, hidden, and output layers. It uses the algorithm of back-propagation to classify the input data. \\[1ex] 
 \hline
\end{tabular}
\setlength{\textfloatsep}{0pt}
        \caption{Classifiers in FER.}
    \label{tab:Classifiers in FER}
\end{table*}
\setlength{\textfloatsep}{0pt}
\section{Audio-Visual Emotion Recognition}

The third emotion recognition application is to recognize emotions from audio-visual input \cite{hossain2019emotion}. The study used deep learning algorithms on big emotional data. The classification method used is support vector machine (SVM). After extraction and classification of data, the output was fed to an extreme learning machine (ELM) as a fusion stage to predict the input features' emotions. Another work in \cite{ouyang2017audio} studied audio-visual emotion recognition by using two distinct state of the art methods. These methods are deep transfer learning and multiple temporal models. They used different DCNN for feature extraction and classification, and their result showed competitive performance. Moreover, the study in \cite{zhang2016multimodal} used DCNN by incorporating it with multimodal systems for recognizing emotions. They tested their method on the RML dataset, and the result showed an improvement in the accuracy.

People interact with each other by using various types of expressions. Emotions are expressed clearly by using verbal and nonverbal communication, and therefore it is easier to understand each other. Expanding the research to include many expressions has the advantage of facilitating research to recognize emotions. Furthermore, by mixing audio and visual expression, researchers can benefit from the big data that will be created because one of the limitations of using either speech or facial is the limited number of datasets.

In Audio-Visual Recognition (AVR), three steps are applied to both speech and facial, then fusion is applied to extract the final emotion expression. These three steps are preprocessing, feature extraction then finally classification.

\subsection{Preprocessing}
Data selection in audio-visual emotion (AVM) recognition is usually taken from videos. 
The videos' content is first portioned into many frames where they are the main source of visual-based features. Videos have the advantage of controlled datasets. These datasets have a fixed setup in which it focuses on the face's area as it contains all the expressions that give the specific emotion\cite{ouyang2017audio}. 
For audio signals, the signals are extracted and converted to 16k Hz sampled by using quantized mono signals. 
The audio features take advantage of an extended version of GeMAPS (eGeMAPS), and the features are normalized to be zero mean and unit variance to be ready to forward to DNN input\cite{li2018mec}. 

The study in \cite{noroozi2017audio} uses preprocessing technique by applying the Multi-Task Cascaded Convolutional Network (MTCNN) to extract the face expression and alignment for video frames. It divides the task into small segments, and CNN trains all the sub-tasks to confidently detect the features and prepare them for classification. Authors in \cite{avots2019audiovisual} extracted frames at the beginning and end of each video for preprocessing the input data. They followed this method to avoid the repetition of training the systems with the same facial expressions.

\subsection{Feature Extraction}
Feature extraction is the first stage to recognize emotions; therefore, it affects the whole system's performance. Recently, DCNN has been used to extract features and prepare them for the next stage. Authors in \cite{zhang2016multimodal} used different methods to extract the features from facial and audio contents. They used prosodic, Mel-frequency Cepstral Coefficient (MFCC) to extract the features from facial and Gabor wavelet from audio, then combine the two extract features to represent audio-visual features and pass them to the final stage, classification.

Authors in \cite{shu2020multimodal} used statistical parts of the audio signal's energy and pitch contours to extract audio features, and they used the faces motion features such as the movements of the head, eyebrows, eyes, and mouth to record the facial features. After that, they send each feature separately to the classification stage to classify them and identify the specific emotion. There are many feature extraction methods to improve the systems performance and in table ~\ref{tab:Feature Extraction in AVR}, we summarized the common three used methods.
\begin{table*}[!ht]
\label{mainref}
    \centering
    \begin{tabular}{|c|p{1.8cm}|p{13.5cm}|}\hline
 Reference & Classifier & Description \\ [0.5ex] 
 \hline
 \cite{ouyang2017audio} & VGG-LSTM & In this feature extraction model, the authors used VGG-16 to receive layers and extract the features from them, and then they will be forwarded to LSTM layers to recognize the emotion.\\ 
 \hline
 \cite{ouyang2017audio} & ResNet-LSTM & In this model, the features are extracted from different layers as sub-tasks and they will be passed to LSTM layers.\\
 \hline
 \cite{ouyang2017audio} & C3D Network & In this model C3D network is the  instead of using traditional 2-D kernels to improve the system performance.\\ [1ex] 
 \hline
\end{tabular}
    \caption{Feature Extraction in AVR.}
    \label{tab:Feature Extraction in AVR}
\end{table*}
\setlength{\textfloatsep}{0pt}
\subsection{Classification and fusion}
The study in \cite{avots2019audiovisual} used AlexNet to do the classification process of multimodal emotion recognition, and they compared it with the model of validation of human decision-makers. They tested it on different datasets such as RML, SAVEE, eNTER. The result of this system is competitive with other studies in the same field.

After classifying both speech and visual, the fusion stage comes to accommodate both layers to deliver the final decision of emotion in multimodal systems. The main function is to make layers that come from the speech section and visual section at the same length. The authors in \cite{ouyang2017audio} used the brutal force technique to find the optimum weight of incoming layers. At this step, the system gives the final decision of incoming expression and recognizes the emotion.

\section{DISCUSSION AND FUTURE WORK}
In this section, we present the challenges and possible future directions. We start by discussing the datasets that are available for applying the different processes. Then, we discuss the three mentioned techniques separately by mentioning their strength and weakness.
\subsection{datasets}
Most of the available datasets are acted expressions, and they are produced at studios or labs. These studios contain high-quality recordings, and they are noise-free. One of the challenges includes choosing a suitable method to extract features, system robustness to tone changes, the talking style, the speaking rate, the cultures and environment of people, which affect the way of expressing emotions. Moreover, most of the acted expressions are created from the same person, making them not real. The effectiveness of these datasets to be applied in real life where there is noise, and people's natural expressions are different from the features imitated in the studios depend on the content of the datasets' content. Therefore, these datasets' effectiveness is challenging, and the accuracy of the classified expressions is questionable. The suggested solution is to use real datasets based on real experiments to be sure of the different techniques.
\subsection{Speech based ER}
The feature selection and extraction method are aimed to prepare the data from utterances and noises and pass them to the classifiers to classify the data. One challenge in SER is that collecting and annotating large utterances are difficult due to the hardness of processing large datasets, especially the speech signals that are continuous by nature. All the available feature selection methods are designed to process small datasets. Even though deep learning-based classifiers are used to classify the input data, it is still challenging to choose a suitable classifier compatible with the used method in feature selection and extraction. Hence selecting a classifier that can improve the system's performance and increase the classification accuracy is an open problem to be discussed in future works.
\subsection{Facial based ER}
The majority of the existing methods that are used in FER are based on different training datasets and training the variations of the expressions from the image such as illuminations, head pose, the distance between the corners of eyes and head, etc. The image is divided into subframes and layers that are processed to select and extract emotions from these layers. Therefore, training deep layers and flexible filters are sufficient for the training process to extracting the image's features and expressions. Nevertheless, this method is sensitive to the trained dataset's size, which might cause degradation in the classification accuracy and the whole classification method's performance. Hence, recently CNN has been used in the classification stage, especially CNN can train deep layers effectively as they are a good solution for the head pose variations and calculating the distances of corner points in the face.
\subsection{Audio-Visual based ER}
In audio-visual-based emotion recognition, most techniques preprocess, select and extract, and classify speech and facial features on their own and then combine them in a fusion stage. The accuracy of the whole system depends on the fusion stage, and the synchronization between features coming from speech and facial features is crucial. Many fusion scenarios have been introduced to enhance the systems' accuracy, such as SVM, PCA, SVM-PCA, etc. Therefore, studying the fusion method that achieves exemplary performance is an open problem and needs looking.  
\section{CONCLUSION}

In this paper, we present a survey of using convolutional neural networks to recognize patients emotions. We started by showing the different databases used in this field, with brief information about their constructions to do emotion recognition. Three important applications of using CNNs in emotion recognition were studied, which are speech-based emotion recognition, facial-based emotion recognition, and audio-visual-based emotion recognition. We studied in detail each section and explained the different approaches they used. The study of each section focuses on pre-processing, feature selection and extraction, and the classification method of each stage to identify the final expressions.


\section*{Acknowledgment}
This work was made possible by NPRP grant \# NPRP13S-0205-200265 from the Qatar National Research Fund (a member of Qatar Foundation). The findings achieved herein are solely the responsibility of the authors.  
\bibliographystyle{IEEEtran}
\bibliography{bibliography.bib}

\end{document}